\documentclass{article}

\PassOptionsToPackage{numbers, compress}{natbib}

\usepackage[final]{neurips_2023}
\usepackage{tcolorbox}
\usepackage{graphicx}

\usepackage[utf8]{inputenc} 
\usepackage[T1]{fontenc}    
\usepackage{hyperref}       
\usepackage{url}            
\usepackage{booktabs}       
\usepackage{amsfonts}       
\usepackage{nicefrac}       
\usepackage{microtype}      
\usepackage{xcolor}         
\usepackage{multirow}
\usepackage{float}
\usepackage{natbib}

\usepackage{comment}
\usepackage{caption}

\title{Deficiency of Large Language Models in Finance: \\ An Empirical Examination of Hallucination}

\begin{document}

\author{Haoqiang Kang$^{1}$\thanks{Haoqiang Kang completed this work as a research assistant at Columbia University.},~~Xiao-Yang Liu$^{1, 2}$\\
$^{1}$Columbia University, \\
$^{2}$Rensselaer Polytechnic Institute\\
\footnotesize{\texttt{haoqik88@gmail.com, XL2427@columbia.edu}}
}

\maketitle

\begin{abstract}

The hallucination issue is recognized as a fundamental deficiency of large language models (LLMs), especially when applied to fields such as finance, education, and law. Despite the growing concerns, there has been a lack of empirical investigation. In this paper, we provide an empirical examination of LLMs' hallucination behaviors in financial tasks. First, we empirically investigate LLM model's ability of explaining financial concepts and terminologies. Second, we assess LLM models' capacity of querying historical stock prices. Third, to alleviate the hallucination issue, we evaluate the efficacy of four practical methods, including few-shot learning, Decoding by Contrasting Layers (DoLa), the Retrieval Augmentation Generation (RAG) method and the prompt-based tool learning method for a function to generate a query command. Finally, our major finding is that off-the-shelf LLMs experience serious hallucination behaviors in financial tasks. Therefore, there is an urgent need to call for research efforts in mitigating LLMs' hallucination.\footnote{We release our code and data at \url{https://github.com/mk322/fin_hallu}.}




\end{abstract}

\section{Introduction}

\textbf{Motivation}. Large language models (LLMs) \cite{brown2020language,OpenAI2023GPT4TR,touvron2023llama} have emerged as transformative tools, demonstrating unprecedented prowess in comprehending and generating human-like text across diverse applications. LLMs are revolutionizing the interface\footnote{There may be a switch from graphical user interface (GUI) to language user interface (LUI).} between humans and machines in fields such as finance, education, and law. Among many LLM applications, the finance domain stands out as a notable area of impact. Various specialized models, such as FinBERT \cite{yang2020finbert}, BloombergGPT \cite{wu2023bloomberggpt}, and FinGPT \cite{liu2023fingpt, yang2023fingpt}, have been customized to understand and generate financial texts, showing considerable promise in aiding humans in a variety of financial tasks, from portfolio management \cite{ko2023can} to predictive analysis of market trends \cite{zhang2023instruct} and sentiment analysis \cite{zhang2023instruct,zhang2023icaif}. However, a fundamental deficiency inherent in these models is \textit{hallucination — generating plausible but unsupported or factually incorrect content to a reference text —} which may be highly risky when deploying financial large language models (FinLLMs). Considering the sensitive and intricate characteristics of finance use cases, inaccuracies and misinformation may lead to severe consequences, such as substantial monetary losses and erosion of trust.

\textbf{Challenges}. 
The hallucination issue is considered as a major challenge of deploying FinLLMs in real-world applications. Addressing the hallucination issue is nontrivial. First, the problem of properly measuring hallucinations, \textit{determining how often and to what extent LLMs produce incorrect or "hallucinated" information}, particularly amidst intricate financial concepts. Second, the challenge of deployment in real-world financial scenarios. Finance is a multifaceted field, while tasks such as querying historical stock prices demand pinpoint accuracy. \textit{Will LLMs be able to consistently deliver on these demands}? 


\textbf{Contributions}. In this paper, taking an empirical approach, we examine hallucination behaviors of LLMs in financial tasks. We summarize our work as follows:
\begin{itemize}
\item \noindent \textbf{Examining LLMs' ability of financial knowledge}:
We conduct an empirical investigation to elucidate the extent of hallucinations in finance, produced by LLMs. Our analysis delves into the models' capacity of memorizing and retaining financial domain knowledge, providing critical insights into their reliability in grasping intricate financial concepts and terminologies.

 \item \noindent\textbf{A case study of analyzing a real-world financial task}:
As a case study, we further analyze the models' performance in a fundamental financial task, namely the ability of querying historical stock prices accurately. This enables us to discern the models’ practical utility and effectiveness in addressing tasks that are quintessential in the finance sector.

\item \noindent \textbf{Evaluating mitigation methods for hallucinations}:
We assess four practical methods, few-shot prompting, Decoding by Contrasting Layers (DoLa)~\cite{chuang2023dola}, the \textit{Retrieval Augmentation Generation (RAG)} method and the \textit{prompt-based tool learning method} that generates correct function calls. These strategies are designed to enhance the factual accuracy of the generated outputs and enable models to interact with up-to-date financial data, ensuring the relevance and reliability of the information provided.
\end{itemize}

These contributions underscore the need to reduce hallucinations of LLMs and enhance their reliability for practical financial tasks in the real world.

\section{Background and Related Works}

\begin{figure*}
    \centering
    \includegraphics[width=1\textwidth]{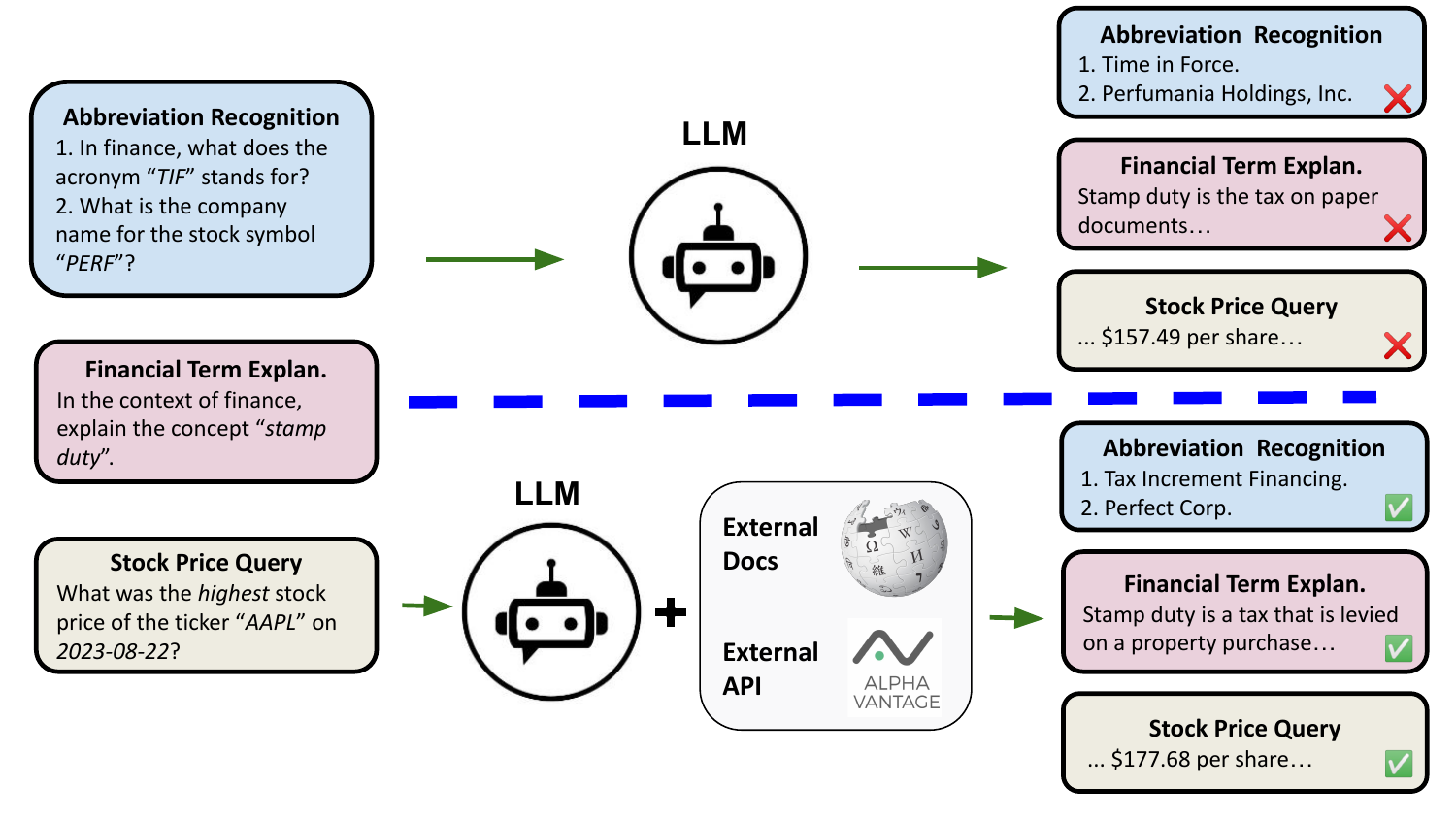}
    \caption{An overview and sample outputs from our empirical analysis of three financial tasks. The sample outputs for the abbreviation recognition task are generated by GPT4, while the others are generated by Llama2-7B-Chat.}
    \label{fig:enter-label}
\end{figure*}

The concept of "hallucination" in the context of LLM refers to instances where the generated text conflicts with either the input instructions (\textit{instruction inconsistency}), the input context \label{lab:def_hallu}(\textit{input context inconsistency}), the previously generated text (\textit{generated context conflict}), or established world knowledge (\textit{factual inconsistency})~\cite{huang2023survey, zhang2023siren}. In this work, we focus particularly on \textit{factual inconsistency}, as it represents a serious and frequent type of error. For instance, as illustrated in Figure~\ref{fig:enter-label}, the GPT4 model incorrectly interprets the financial acronym "TIF" as "Time in Force", which is not the common full name for "TIF", instead of the correct "Tax Increment Financing."


\textbf{Hallucinations in general-purpose LLMs}. Previous research has investigated several factors contributing to LLM hallucination, such as imperfect learning and decoding methods, and knowledge gaps in the training data~\cite{ji2023survey}. In addition, various methods have been proposed to mitigate hallucination in LLMs. One line of work involves factuality-enhanced decoding techniques~\cite{chuang2023dola, shi2023trusting, li2022Contrastive, li2023inference}. Other works leverage external tools, notably the RAG technique, which enhances factuality by incorporating external knowledge sources in the generation process \cite{lewis2020retrieval, jiang2023active, asai2023self}, and the application of precise API calls~\cite{patil2023gorilla, schick2023toolformer} to obtain up-to-dated knowledge. In this study, we examine the effectiveness of one of the most representative decoding method of DoLa~\cite{chuang2023dola} method, a traditional RAG approach~\cite{lewis2020retrieval}, and an prompt-based tool learning method in mitigating hallucinations within the finance domain.
 

\textbf{Hallucinations in specific domains}. The deployment of LLMs in specialized domains faces a significant challenge due to the risk of hallucinations. This is particularly critical in areas like finance, medicine and law where accuracy is paramount. Research in this area is growing, with notable efforts like the Med-HALT benchmark by~\citet{umapathi2023med} assessing hallucinations in the medical domain. Similarly, a study of ChatGPT~\cite{wang2023performance} in the Chinese medical sector demonstrates GPT4's advancements, yet also underscores the ongoing challenges. In the legal field, the ChatLaw model~\cite{cui2023chatlaw} employs a vector database and keyword retrieval to tackle hallucinations in legal data extraction. Additionally, \citet{elaraby2023halo} advocate for incorporating domain-specific knowledge, such as NBA data, into supervised-finetuned (SFT) datasets to reduce such inaccuracies.

\section{Methodology for Empirical Examination}

We introduce an empirical framework tailored for evaluating the hallucination behaviours of LLMs within the financial field, as shown in Fig.~\ref{fig:enter-label}. The framework presents a method to assess LLM responses on three financial tasks: recognizing financial abbreviations, explaining financial terminologies, and fetching stock prices. When provided with standalone questions, the models sometimes hallucinate with wrong facts, but the performance significantly improves when integrated with external data sources, such as financial documents and a function to generate a query command.

\subsection{Large Language Models (LLMs)}

Throughout our experiments, we use the HuggingFace weights\footnote{https://huggingface.co/meta-llama/Llama-2-7b-chat-hf}\footnote{https://huggingface.co/meta-llama/Llama-2-7b-hf} of the pretrained Llama2-7B model \cite{touvron2023llama2} and its instruction-tuned+RLHF version Llama2-7B-chat \cite{touvron2023llama2}. Also, we utilize the OpenAI API to call the models of GPT3.5-turbo\footnote{https://openai.com/blog/chatgpt}, and GPT4\footnote
{https://openai.com/research/GPT4}. We use the greedy decoding method to generate texts for all models. In addition to the general-purpose LMs, we investigate the performance of FinMA-7B-NLP~\cite{xie2023pixiu}, a multi-task fine-tuned LLaMA-1-7B model~\cite{touvron2023llama} with instruction data on five finance tasks, including question answering.\footnote{To optimize FinMA's ability to follow instructions effectively, we employ the same prompt template that the original authors used during the finetuning process.} This evaluation includes a comparison with its base model, LLaMA-1-7B~\cite{touvron2023llama}, to highlight differences on hallucination after domain-specific finetuning.

\textbf{Budget}. For the inference process with the Llama2-7B and Llama2-7B-chat models, we utilized an A100 GPU. The cumulative GPU time for our entire project amounted to roughly 40 hours. Regarding the generation tasks with the GPT3.5-turbo and GPT4 models, the associated API costs were approximately \$200. Moreover, the API costs for employing FactScore with GPT3.5-turbo were approximately \$250.

\subsection{Financial Tasks}
When selecting representative tasks in finance, we considered factors such as their relevance to real-world financial activities and the potential risks of inaccurate outputs. Based on these considerations, we identified three primary tasks, which together provide an assessment of LLM performance. Examples of prompt and outputs for these three tasks are gave in Appendix~\ref{app:ex-output}.

\subsubsection{Task I: Financial Abbreviation Recognition}

\textbf{Task description}.
In this task, we measure a LLM model's ability to recognize financial acronyms and stock symbols. We randomly select $192$ financial acronyms from Wikipedia \footnote{https://en.wikipedia.org/wiki/List\_of\_business\_and\_finance\_abbreviations} and $1215$ stock symbols from an online list\footnote{https://eoddata.com/symbols.aspx. We randomly select symbols that are in the stock exchanges of NASDAQ , NYSE, and AMEX.}. This selection is designed to cover a wide range of financial contexts and complexities. The primary objective is to evaluate a LLM mode's capability to accurately expand financial acronyms or provide the full company names corresponding to specific stock symbols. See Table~\ref{tab:app_abbr_examples} and Table~\ref{tab:demo_abbr} in the Appendix for example outputs and the prompt template used in this task.


\textbf{Metric}. The accuracy for both tasks is measured using a consistent approach. The accuracy score is calculated as the ratio of accurately identified acronyms to the total number, which is 192 or 1000, depending on the task. An example is considered to be \textit{correctly recognized} if the actual full name's string is a substring of the predicted full name.


\subsubsection{Task II: Financial Term Explanations}

\textbf{Task description}. In this task, a LLM is prompted to give an explanation of a financial terminology. We randomly select 160 infrequently-visited financial terminologies from Wikidata API\footnote{https://query.wikidata.org/}, specifically targeting those with the lowest page views between 2021-01-01 and 2023-01-01. This selection process aim to emphasize more obscure financial concepts that are less commonly encountered in typical discussions. See Table~\ref{tab:app_term} and Table~\ref{tab:demo_long-form} in the Appendix for example outputs and the prompt templates used in this task.


\textbf{Metric}. We employ the FactScore (ChatGPT+Retrieval) metric \cite{min2023factscore} to measure the factuality of our generated content. This method quantifies the ratio of the number of correct atomic facts to the total number of atomic facts within a given response, benchmarking against the content of each term's explanation (on the Wikipedia full page).

\subsubsection{Task III: Stock Price Query}

\textbf{Task description}. In this task, we query LLMs for a historical stock price with 560 examples in total. We randomly select 70 stock tickers. For these tickers, we determine a set of four dates that is in the period covered by Llama2's pretraining data (i.e., before September 2022) \cite{touvron2023llama2}, specifically on 2022-05-23, 2022-06-22, 2022-07-22, and 2022-08-22. See Table~\ref{tab:app_price} and Table~\ref{tab:demo_price} in the Appendix for example outputs and the prompt templates used in this task.


\textbf{Metric}. In scenarios where no external tool is utilized, our assessment metrics include 1) accuracy, which is the percentage of the predicted prices that are exactly the same to their corresponding actual prices \footnote{Obtained from the Alpha Vantage API:  https://www.alphavantage.co/}, 2) mean absolute error (MAE), and 3) the percentage of queries where the integer parts of the predicted values are the same as that of the actual price. In the prompt-based tool learning scenario, a response is considered to be correct only if the generated query code of a function call exactly matches one of the predefined set of expected function calls. 


\subsection{Methods for Mitigating Hallucination}

\textbf{Few-shot prompting}. We hand crafted the few-shot prompts for each task, utilizing them to guide the models in their learning process.

\textbf{Decoding by contrasting layers} (DoLa). DoLa~\cite{chuang2023dola} is designed to enhance the factual accuracy of LLMs by contrasting the outputs from different layers of the model. This approach assumes that higher layers of the model contain more factual knowledge, which can be leveraged to reduce hallucinations and improve the accuracy of response. In our implementation of DoLa, we utilized the same contrasting layer configurations as those specified by the original authors\footnote{https://github.com/voidism/DoLa}.

\textbf{Retrieval augmentation generation} (RAG). In the tasks of acronym recognition and long-form generation of explanation of financial terms, to further improve the relevance and factuality of the generated content, we integrate RAG \cite{lewis2020retrieval} that sources content directly from Wikipedia. By leveraging the vast informational expanse of Wikipedia, we ensure that our generated outputs are grounded in factual and current knowledge. To make this retrieval process efficient and accurate, we employ the FAISS vector store \cite{johnson2019billion}, which ensures that the most relevant content is extracted from Wikipedia and seamlessly incorporated into our model's outputs.

\textbf{Prompt-based tool learning}. For the stock price query task, our aim is to equip the model with the capability to use external tools. We employ a prompt-based learning technique. In each query, we assess the model's ability in generating the correct Python function call for a tailored wrapper of the Alpha Vantage API based on given natural language instructions. The model is briefed about the function parameters, encompassing the ticker, date, and price type. With this information, the model must produce the function name and the parameters correctly, adhering to Python's syntax. A response is considered as correct only if it precisely matches a set of the expected query strings.

\section{Empirical Results for Hallucination}

\subsection{Quantifying Hallucinations}

Our main results across the three benchmark tasks are given in Table \ref{tab:reference} and Table \ref{tab:stockprice_whole}, and our main findings are as follows.

\textbf{General-purpose LLMs generate factually incorrect content in finance}. An initial evaluation of relatively smaller, open-source models, specifically Llama2-7B and Llama2-7B-chat, reveals their comparatively limited capacity to assimilate extensive knowledge within the finance domain. Delving into more sophisticated models, an examination of the GPT4 model, as presented in Table \ref{tab:reference}, demonstrates a 82.5\% and 90.4\% accuracy when we directly query it in the acronym and stock symbol test. Also, in the long-form generation task that the employed FactScore metric \cite{min2023factscore} delineates an 81.11\% score for GPT4 in a direct query setting. While these performances are notable, they inherently suggest a room for error. Taking a deeper look, upon examining the incorrect responses, we notice that certain generated answers are outdated, stemming from obsolete information. For instance, as illustrated in Figure~\ref{fig:enter-label}, the GPT4 model incorrectly references "PERF" as the stock symbol for "Perfumania Holdings," not accounting for its delisting, highlighting a lapse in updating its knowledge base. In the sensitive and dynamic finance domains, these deviations and outdatedness can manifest into significant consequences. Consequently, it is essential for researchers and professionals to adopt a discerning approach when utilizing outputs from large language models and emphasize the necessity of autonomous validation for pivotal data they generate.

\textbf{Multi-task domain-specific finetuning could diminish LMs' general instruction-following abilities.} As demonstrated in Table~\ref{tab:reference}, FinMA-7B—a model fine-tuned for specific tasks within the finance domain—underperforms its base model, Llama1-7B, in various tasks under both zero-shot and few-shot settings. This trend indicates that while multi-task domain-specific finetuning aims to bolster a model's domain-specific capabilities, it might also lead to a decrease in its overall ability to accurately follow instructions and adapt to new tasks. Such a decrease could result in more frequent occurrences of instruction-inconsistent hallucinations~\ref{lab:def_hallu}. This observation serves as a cautionary note for the utilization of multi-task finetuning in the development of future domain-specific language models.


\textbf{General-purpose LLMs generate seriously unreliable real-world financial predictions}.  The application of LLMs to real-world tasks in the finance domain, particularly in predicting stock prices, raises significant concerns regarding the reliability of their outputs, as highlighted by the results presented in Table \ref{tab:stockprice_whole}. In the zero-shot setup, models such as Llama2-7B and Llama2-7B-chat exhibit alarmingly high Mean Absolute Errors (MAE) of 6357.6 USD and 6380.5 USD, respectively. Furthermore, while the utilization of few-shot prompting allows models to generate responses that are closer to correct answers, there still exists a gap in terms of MAE and accuracy. This high deviation from the true values suggests substantial inaccuracies in their predictions.

\begin{table*}
        \centering
        \caption{Results of the acronym recognition task and terminology explanation task. The GPT-3.5 Turbo and GPT-4 models abstain from answering 56.5\% and 42.3\% of questions, respectively. Their accuracy are evaluated only on the subset of questions to which they provide answers.}
        \begin{tabular}{lcccc}
            \toprule
            &  &\multicolumn{1}{c}{Acronym} &\multicolumn{1}{c}{Stock Symbol} & \multicolumn{1}{c}{Term explanation} \\
            Model  & Method    & Accuracy (\%) & Accuracy (\%)  & FactScore (\%) \\
            \midrule
            Llama1-7B     & Zero-Shot & 40.5 & 14.3 & 46.34\\
            Llama2-7B     & Zero-Shot & 50.1 & 16.0 & 51.94\\
            Llama2-7B-chat& Zero-Shot  & 74.8 & 16.5  & 64.68\\
            Llama2-13B     & Zero-Shot & 58.3 & 24.2  & 56.08  \\
            Llama2-13B-chat& Zero-Shot  & 75.0 & 12.2 & 66.72 \\
            GPT3.5-turbo  & Zero-Shot  & 76.8 & 87.7$^*$ & 78.54\\
            GPT4           & Zero-Shot & 82.5 & 90.4$^*$   & 81.11\\
            FinMA-7B & Zero-Shot  & 30.4 & 7.4   & 25.56 \\
            \midrule
            Llama1-7B     & Few-Shot & 54.3 & 22.7 & 49.27\\
            Llama2-7B     & Few-Shot & 57.3 & 23.0 & 51.60\\
            Llama2-7B-chat& Few-Shot  & 77.1 & 20.0  & 67.02\\
            Llama2-13B     & Few-Shot & 62.7 & 25.4  & 63.28\\
            Llama2-13B-chat& Few-Shot  & 76.1 & 23.7 & 67.97\\
            FinMA-7B & Few-Shot  &  36.5 & 9.8   & 27.64 \\
            \midrule
            Llama2-7B    & DoLa & 41.5 & 17.3 & 56.39 \\
            Llama2-7B-chat& DoLa & 61.8& 17.6 & 65.42 \\ 
            Llama2-13B    & DoLa & 42.8 & 23.4 & 67.06 \\
            Llama2-13B-chat& DoLa & 62.4 & 13.7  & 67.50 \\ 

            \midrule
            Llama2-7B     & RAG   & 86.6 & 61.5   & 75.07\\
            Llama2-7B-chat& RAG   & 90.8 & \textbf{69.3}  &  90.48\\
            Llama2-13B     & RAG   & 87.4 & 62.8& 78.56\\
            Llama2-13B-chat& RAG   & \textbf{93.4}& 68.5 & \textbf{90.67}  \\

            \bottomrule
        \end{tabular}
        \label{tab:reference}
\end{table*}

Interestingly, the models of GPT3.5-turbo and GPT4 opt for a more cautious approach, abstaining from generating responses to some of stock symbol recognition questions and all stock-price-related questions in the absence of external tools. This restraint is praiseworthy as it prevents the propagation of potentially erroneous and misleading financial predictions to users, fostering user trust and minimizing the risk of misinformation in such a critical domain.

\subsection{Mitigating Hallucination}

\begin{table}[ht!]
    \centering
    \caption{The comparative performance  of different models for the stock price query task, with and without an external tool. $^{*}$The OpenAI models abstain to answer any question related to stock prices.}
    \begin{tabular}{lcccc}
        \toprule
        Models               & Valid (\%) $\uparrow$ & MAE (\$) $\downarrow$& int\_correct (\%) $\uparrow$& Accuracy (\%) $\uparrow$\\
        \midrule

        \midrule
        \multicolumn{5}{c}{\textit{Zero-Shot}} \\
        Llama2-7B           & 85.3     & 6357.6  & 2.0            & 0.0 \\
        Llama2-7B-chat      & 100.0     & 6380.5 & 2.7             & 0.0\\
        GPT3.5-turbo$^{*}$  & 0.0         & -        & -      & - \\
        GPT4$^{*}$  & 0.0   & -        & -      & - \\

        Llama2-7B\textcolor{purple}{+DoLA}           &76.8    & 6857.9  & 0.9           & 0.0 \\
        Llama2-7B-chat\textcolor{purple}{+DoLA}      & 97.4     & 6460.2 & 1.2            & 0.0\\
        Llama2-7B\textcolor{blue}{+tool}     & 47.5    & -        & -                 & 76.4 \\
        Llama2-7B-chat\textcolor{blue}{+tool}& 89.8    & -        & -                 & 91.2 \\

        GPT3.5-turbo\textcolor{blue}{+tool}  & 100.0         & -        & -                 & 98.4 \\
        GPT4\textcolor{blue}{+tool}   & 100.0          & -        & -                 & 100.0 \\
        \midrule
        \multicolumn{5}{c}{\textit{One-Shot}} \\
        Llama2-7B           & 90.2    & 489.3  & 2.1           & 0.0 \\
        Llama2-7B-chat      & 100.0     & 234.5 & 3.4            & 0.0\\
        Llama2-7B\textcolor{purple}{+DoLA}           &86.6    & 521.6  & 2.1          & 0.0 \\
        Llama2-7B-chat\textcolor{purple}{+DoLA}      & 97.4     & 234.9 & 2.4            & 0.0\\
        Llama2-7B\textcolor{blue}{+tool}     & 97.4     & -        & -                 & 100.0 \\
        Llama2-7B-chat\textcolor{blue}{+tool}& 99.8      & -        & -                 & 100.0 \\

        GPT3.5-turbo\textcolor{blue}{+tool}  & 100.0         & -        & -                 & 100.0 \\
        GPT4\textcolor{blue}{+tool}   & 100.0        & -        & -                 & 100.0 \\

        \bottomrule
    \end{tabular}
    
    \label{tab:stockprice_whole}
\end{table}

\textbf{RAG significantly improves the factuality in finance}. Evaluating the impact of the RAG on both foundational and finetuned models provides compelling evidence of its efficacy. As can be seen in Table \ref{tab:reference}, integrating RAG consistently elevates the performance of both Llama-2 and Llama-2-chat models. Table \ref{tab:reference} further underscores the advantage of RAG, showing substantially higher FactScores in the long-form generation task for both models when RAG is implemented. Such consistent improvements across multiple metrics and settings confirm that RAG serves as a significant enhancement to both pretrained and instruction-tuned models.

\textbf{Prompt-based tool learning helps significantly on the time-sensitive task}.
As discerned from Table \ref{tab:stockprice_whole}, the smaller models, Llama2-7B and Llama2-7B-chat, without the integration of an external tool, exhibit negligible accuracy in handling stock queries. Conversely, the application of the prompt-based tool learning leads to a transformative elevation in performance, with Llama2-7B+tool and Llama2-7B-chat+tool achieving remarkable accuracies of 100.00\% with only one training example. This enhancement in reliability and precision is critical, especially within the rapidly evolving landscape of financial domains. 
It's notable that the sophisticated models, GPT3.5-turbo and GPT4, refrain from addressing stock price queries in a zero-shot setting. However, when augmented with prompt-based tool tool learning, these models are adeptly optimized to provide accurate answers. The integration of zero-shot and few-shot tool learning thus emerges as an effective strategy bridging the knowledge gap and enhancing the reliability of language models, especially in the dynamic domain of finance.

\textbf{Few-shot learning better improves the ability to follow the question-answering format than factuality}. As shown in Table~\ref{tab:reference}, we observe that the pretrained Llama2-7B and Llama2-13B models improve significantly in their question-answering capabilities under few-shot learning, compared to their zero-shot performance. However, this improvement does not markedly surpass the performance of their zero-shot, instruction-tuned counterparts, Llama2-7B-chat and Llama2-13B-chat. Furthermore, for the chat variants, few-shot learning's impact is more limited, yielding only modest improvements. This trend suggests that while few-shot learning aids in adapting to question formats, its role in enhancing factual precision is less substantial.

\textbf{DoLa has limitations in enhancing models with knowledge gaps in training data}. The DoLa decoding method is designed to enhance the factual accuracy of LMs by contrasting the outputs from different layers of the model. This approach assumes that higher layers of the model contain more factual knowledge, which can be leveraged to reduce hallucinations and improve the accuracy of responses. However, the effectiveness of DoLa is inherently dependent on the breadth and depth of knowledge encoded in the language model's pretraining dataset. As shown in Table~\ref{tab:reference} and Table~\ref{tab:stockprice_whole}, while DoLa can improve factual accuracy in some instances, its effectiveness is limited when the underlying model lacks comprehensive knowledge in its pretraining dataset, such as some stock ticker and stock prices. This limitation is particularly noticeable in scenarios where the model is expected to provide responses based on information that might not be well-represented or current in its training data. Since DoLa relies on amplifying the knowledge already present in the model, its ability to compensate for gaps in the model's foundational knowledge is constrained.

\section{Conclusion, Limitations and Future Works}

In this study, we conduct an empirical analysis to assess the hallucination problem of LLMs in the financial domain. Our work demystifies the LLM's reliability and ability of explaining financial terminologies and concepts. Furthermore, a performance analysis reveals the practical viability and performance of these models in querying the historical stock prices. To mitigate hallucinations, we show the effectiveness of the RAG method and prompt-based tool learning to generate correct function calls, thereby ensuring the provision of factually correct and up-to-date information in the finance domain.

While our research provides insights into the capabilities of LLMs in the finance domain, it is essential to acknowledge certain limitations. Firstly, our tasks, although representative, cannot encompass the full spectrum of real-world tasks in the vast and varied domain of finance. This implies that our results may not generalize to all possible financial tasks and scenarios. Secondly, the mitigation strategies we introduced and tested are task-specific. Their effectiveness might vary when applied to different tasks or in diverse financial contexts. Future research might need to adapt or expand these strategies to ensure their relevance and effectiveness across a broader range of financial applications.

Moving forward, potential avenues for future works include refining the hallucination mitigation techniques for broader financial applications, exploring ways to further increase the accuracy and reliability of LLMs in dynamic financial settings, and understanding the interplay between LLM outputs and financial decision-making. Our findings highlight the critical issue of hallucinations, establishing a groundwork for advancing responsible and reliable LLM deployment in the financial domain.

\bibliographystyle{plainnat}
\bibliography{custom}

\begin{thebibliography}{30}
\providecommand{\natexlab}[1]{#1}
\providecommand{\url}[1]{\texttt{#1}}
\expandafter\ifx\csname urlstyle\endcsname\relax
  \providecommand{\doi}[1]{doi: #1}\else
  \providecommand{\doi}{doi: \begingroup \urlstyle{rm}\Url}\fi

\bibitem[Asai et~al.(2023)Asai, Wu, Wang, Sil, and Hajishirzi]{asai2023self}
Akari Asai, Zeqiu Wu, Yizhong Wang, Avirup Sil, and Hannaneh Hajishirzi.
\newblock Self-rag: Learning to retrieve, generate, and critique through self-reflection.
\newblock \emph{arXiv preprint arXiv:2310.11511}, 2023.

\bibitem[Brown et~al.(2020)Brown, Mann, Ryder, Subbiah, Kaplan, Dhariwal, Neelakantan, Shyam, Sastry, Askell, et~al.]{brown2020language}
Tom Brown, Benjamin Mann, Nick Ryder, Melanie Subbiah, Jared~D Kaplan, Prafulla Dhariwal, Arvind Neelakantan, Pranav Shyam, Girish Sastry, Amanda Askell, et~al.
\newblock Language models are few-shot learners.
\newblock \emph{Advances in Neural Information Processing Systems}, 33:\penalty0 1877--1901, 2020.

\bibitem[Chuang et~al.(2023)Chuang, Xie, Luo, Kim, Glass, and He]{chuang2023dola}
Yung-Sung Chuang, Yujia Xie, Hongyin Luo, Yoon Kim, James Glass, and Pengcheng He.
\newblock Dola: Decoding by contrasting layers improves factuality in large language models.
\newblock \emph{arXiv preprint arXiv:2309.03883}, 2023.

\bibitem[Cui et~al.(2023)Cui, Li, Yan, Chen, and Yuan]{cui2023chatlaw}
Jiaxi Cui, Zongjian Li, Yang Yan, Bohua Chen, and Li~Yuan.
\newblock Chatlaw: Open-source legal large language model with integrated external knowledge bases.
\newblock \emph{arXiv preprint arXiv:2306.16092}, 2023.

\bibitem[Elaraby et~al.(2023)Elaraby, Lu, Dunn, Zhang, Wang, and Liu]{elaraby2023halo}
Mohamed Elaraby, Mengyin Lu, Jacob Dunn, Xueying Zhang, Yu~Wang, and Shizhu Liu.
\newblock Halo: Estimation and reduction of hallucinations in open-source weak large language models.
\newblock \emph{arXiv preprint arXiv:2308.11764}, 2023.

\bibitem[Huang et~al.(2023)Huang, Yu, Ma, Zhong, Feng, Wang, Chen, Peng, Feng, Qin, et~al.]{huang2023survey}
Lei Huang, Weijiang Yu, Weitao Ma, Weihong Zhong, Zhangyin Feng, Haotian Wang, Qianglong Chen, Weihua Peng, Xiaocheng Feng, Bing Qin, et~al.
\newblock A survey on hallucination in large language models: Principles, taxonomy, challenges, and open questions.
\newblock \emph{arXiv preprint arXiv:2311.05232}, 2023.

\bibitem[Ji et~al.(2023)Ji, Lee, Frieske, Yu, Su, Xu, Ishii, Bang, Madotto, and Fung]{ji2023survey}
Ziwei Ji, Nayeon Lee, Rita Frieske, Tiezheng Yu, Dan Su, Yan Xu, Etsuko Ishii, Ye~Jin Bang, Andrea Madotto, and Pascale Fung.
\newblock Survey of hallucination in natural language generation.
\newblock \emph{ACM Computing Surveys}, 55\penalty0 (12):\penalty0 1--38, 2023.

\bibitem[Jiang et~al.(2023)Jiang, Xu, Gao, Sun, Liu, Dwivedi-Yu, Yang, Callan, and Neubig]{jiang2023active}
Zhengbao Jiang, Frank~F Xu, Luyu Gao, Zhiqing Sun, Qian Liu, Jane Dwivedi-Yu, Yiming Yang, Jamie Callan, and Graham Neubig.
\newblock Active retrieval augmented generation.
\newblock \emph{arXiv preprint arXiv:2305.06983}, 2023.

\bibitem[Johnson et~al.(2019)Johnson, Douze, and J{\'e}gou]{johnson2019billion}
Jeff Johnson, Matthijs Douze, and Herv{\'e} J{\'e}gou.
\newblock Billion-scale similarity search with gpus.
\newblock \emph{IEEE Transactions on Big Data}, 7\penalty0 (3):\penalty0 535--547, 2019.

\bibitem[Ko and Lee(2023)]{ko2023can}
Hyungjin Ko and Jaewook Lee.
\newblock Can chatgpt improve investment decision? from a portfolio management perspective.
\newblock \emph{From a Portfolio Management Perspective}, 2023.

\bibitem[Lewis et~al.(2020)Lewis, Perez, Piktus, Petroni, Karpukhin, Goyal, K{\"u}ttler, Lewis, Yih, Rockt{\"a}schel, et~al.]{lewis2020retrieval}
Patrick Lewis, Ethan Perez, Aleksandra Piktus, Fabio Petroni, Vladimir Karpukhin, Naman Goyal, Heinrich K{\"u}ttler, Mike Lewis, Wen-tau Yih, Tim Rockt{\"a}schel, et~al.
\newblock Retrieval-augmented generation for knowledge-intensive nlp tasks.
\newblock \emph{Advances in Neural Information Processing Systems}, 33:\penalty0 9459--9474, 2020.

\bibitem[Li et~al.(2023)Li, Patel, Vi{\'e}gas, Pfister, and Wattenberg]{li2023inference}
Kenneth Li, Oam Patel, Fernanda Vi{\'e}gas, Hanspeter Pfister, and Martin Wattenberg.
\newblock Inference-time intervention: Eliciting truthful answers from a language model.
\newblock \emph{arXiv preprint arXiv:2306.03341}, 2023.

\bibitem[Li et~al.(2022)Li, Holtzman, Fried, Liang, Eisner, Hashimoto, Zettlemoyer, and Lewis]{li2022Contrastive}
Xiang~Lisa Li, Ari Holtzman, Daniel Fried, Percy Liang, Jason Eisner, Tatsunori Hashimoto, Luke Zettlemoyer, and Mike Lewis.
\newblock Contrastive decoding: Open-ended text generation as optimization.
\newblock \emph{arXiv preprint arXiv:2210.15097}, 2022.

\bibitem[Liu et~al.(2023)Liu, Wang, and Zha]{liu2023fingpt}
Xiao-Yang Liu, Guoxuan Wang, and Daochen Zha.
\newblock {FinGPT}: Democratizing internet-scale data for financial large language models.
\newblock \emph{Workshop on Instruction Tuning and Instruction Following at NeurIPS}, 2023.

\bibitem[Min et~al.(2023)Min, Krishna, Lyu, Lewis, Yih, Koh, Iyyer, Zettlemoyer, and Hajishirzi]{min2023factscore}
Sewon Min, Kalpesh Krishna, Xinxi Lyu, Mike Lewis, Wen-tau Yih, Pang~Wei Koh, Mohit Iyyer, Luke Zettlemoyer, and Hannaneh Hajishirzi.
\newblock Factscore: Fine-grained atomic evaluation of factual precision in long form text generation.
\newblock \emph{arXiv preprint arXiv:2305.14251}, 2023.

\bibitem[OpenAI(2023)]{OpenAI2023GPT4TR}
OpenAI.
\newblock {GPT-4} technical report.
\newblock \emph{ArXiv}, abs/2303.08774, 2023.

\bibitem[Patil et~al.(2023)Patil, Zhang, Wang, and Gonzalez]{patil2023gorilla}
Shishir~G Patil, Tianjun Zhang, Xin Wang, and Joseph~E Gonzalez.
\newblock Gorilla: Large language model connected with massive apis.
\newblock \emph{arXiv preprint arXiv:2305.15334}, 2023.

\bibitem[Schick et~al.(2023)Schick, Dwivedi-Yu, Dess{\`\i}, Raileanu, Lomeli, Zettlemoyer, Cancedda, and Scialom]{schick2023toolformer}
Timo Schick, Jane Dwivedi-Yu, Roberto Dess{\`\i}, Roberta Raileanu, Maria Lomeli, Luke Zettlemoyer, Nicola Cancedda, and Thomas Scialom.
\newblock Toolformer: Language models can teach themselves to use tools.
\newblock \emph{arXiv preprint arXiv:2302.04761}, 2023.

\bibitem[Shi et~al.(2023)Shi, Han, Lewis, Tsvetkov, Zettlemoyer, and Yih]{shi2023trusting}
Weijia Shi, Xiaochuang Han, Mike Lewis, Yulia Tsvetkov, Luke Zettlemoyer, and Scott Wen-tau Yih.
\newblock Trusting your evidence: Hallucinate less with context-aware decoding.
\newblock \emph{arXiv preprint arXiv:2305.14739}, 2023.

\bibitem[Touvron et~al.(2023{\natexlab{a}})Touvron, Lavril, Izacard, Martinet, Lachaux, Lacroix, Rozi{\`e}re, Goyal, Hambro, Azhar, et~al.]{touvron2023llama}
Hugo Touvron, Thibaut Lavril, Gautier Izacard, Xavier Martinet, Marie-Anne Lachaux, Timoth{\'e}e Lacroix, Baptiste Rozi{\`e}re, Naman Goyal, Eric Hambro, Faisal Azhar, et~al.
\newblock Llama: Open and efficient foundation language models.
\newblock \emph{arXiv preprint arXiv:2302.13971}, 2023{\natexlab{a}}.

\bibitem[Touvron et~al.(2023{\natexlab{b}})Touvron, Martin, Stone, Albert, Almahairi, Babaei, Bashlykov, Batra, Bhargava, Bhosale, et~al.]{touvron2023llama2}
Hugo Touvron, Louis Martin, Kevin Stone, Peter Albert, Amjad Almahairi, Yasmine Babaei, Nikolay Bashlykov, Soumya Batra, Prajjwal Bhargava, Shruti Bhosale, et~al.
\newblock Llama 2: Open foundation and fine-tuned chat models.
\newblock \emph{arXiv preprint arXiv:2307.09288}, 2023{\natexlab{b}}.

\bibitem[Umapathi et~al.(2023)Umapathi, Pal, and Sankarasubbu]{umapathi2023med}
Logesh~Kumar Umapathi, Ankit Pal, and Malaikannan Sankarasubbu.
\newblock Med-halt: Medical domain hallucination test for large language models.
\newblock \emph{arXiv preprint arXiv:2307.15343}, 2023.

\bibitem[Wang et~al.(2023)Wang, Wu, Dou, He, and Yang]{wang2023performance}
Hongyan Wang, WeiZhen Wu, Zhi Dou, Liangliang He, and Liqiang Yang.
\newblock Performance and exploration of chatgpt in medical examination, records and education in chinese: Pave the way for medical ai.
\newblock \emph{International Journal of Medical Informatics}, 177:\penalty0 105173, 2023.

\bibitem[Wu et~al.(2023)Wu, Irsoy, Lu, Dabravolski, Dredze, Gehrmann, Kambadur, Rosenberg, and Mann]{wu2023bloomberggpt}
Shijie Wu, Ozan Irsoy, Steven Lu, Vadim Dabravolski, Mark Dredze, Sebastian Gehrmann, Prabhanjan Kambadur, David Rosenberg, and Gideon Mann.
\newblock Bloomberggpt: A large language model for finance.
\newblock \emph{arXiv preprint arXiv:2303.17564}, 2023.

\bibitem[Xie et~al.(2023)Xie, Han, Zhang, Lai, Peng, Lopez-Lira, and Huang]{xie2023pixiu}
Qianqian Xie, Weiguang Han, Xiao Zhang, Yanzhao Lai, Min Peng, Alejandro Lopez-Lira, and Jimin Huang.
\newblock Pixiu: A large language model, instruction data and evaluation benchmark for finance.
\newblock \emph{arXiv preprint arXiv:2306.05443}, 2023.

\bibitem[Yang et~al.(2023)Yang, Liu, and Wang]{yang2023fingpt}
Hongyang Yang, Xiao-Yang Liu, and Christina~Dan Wang.
\newblock {FinGPT}: Open-source financial large language models.
\newblock \emph{Symposium on FinLLM, IJCAI}, 2023.

\bibitem[Yang et~al.(2020)Yang, Uy, and Huang]{yang2020finbert}
Yi~Yang, Mark Christopher~Siy Uy, and Allen Huang.
\newblock Finbert: A pretrained language model for financial communications.
\newblock \emph{arXiv preprint arXiv:2006.08097}, 2020.

\bibitem[Zhang et~al.(2023{\natexlab{a}})Zhang, Yang, and Liu]{zhang2023instruct}
Boyu Zhang, Hongyang Yang, and Xiao-Yang Liu.
\newblock {Instruct-FinGPT}: Financial sentiment analysis by instruction tuning of general-purpose large language models.
\newblock \emph{Symposium on FinLLM at IJCAI}, 2023{\natexlab{a}}.

\bibitem[Zhang et~al.(Nov., 2023)Zhang, Yang, Zhou, Babar, and Liu]{zhang2023icaif}
Boyu Zhang, Hongyang Yang, Tianyu Zhou, Ali Babar, and Xiao-Yang Liu.
\newblock Enhancing financial sentiment analysis via retrieval augmented large language models.
\newblock \emph{ACM ICAIF}, Nov., 2023.

\bibitem[Zhang et~al.(2023{\natexlab{b}})Zhang, Li, Cui, Cai, Liu, Fu, Huang, Zhao, Zhang, Chen, et~al.]{zhang2023siren}
Yue Zhang, Yafu Li, Leyang Cui, Deng Cai, Lemao Liu, Tingchen Fu, Xinting Huang, Enbo Zhao, Yu~Zhang, Yulong Chen, et~al.
\newblock Siren's song in the ai ocean: A survey on hallucination in large language models.
\newblock \emph{arXiv preprint arXiv:2309.01219}, 2023{\natexlab{b}}.

\end{thebibliography}

\appendix

\clearpage
\section{Output Examples}
\label{app:ex-output}

\begin{table}[h]
    \centering
    \begin{tcolorbox}
        \begin{center}
            Examples of Correct Output
        \end{center}
        \textbf{Llama 2 7B}: \\
        \textbf{Prompt}: In finance, the acronym "MICR" stands for \\
        \textbf{Generation}: Magnetic Ink Character Recognition. \\
        \textbf{Prompt}: In finance, the stock symbol "AAPL" stands for \\
        \textbf{Generation}: Apple Inc.\\
        
        \textbf{Llama 2 7B Chat}: \\
        \textbf{Prompt}: In finance, what does the acronym "MICR" stand for? \\
        \textbf{Generation}: MICR stands for Magnetic Ink Character Recognition. It is a technology ... \\
        \textbf{Prompt}: In finance, what does the stock ticker "AAPL" stand for? \\
        \textbf{Generation}: It stands for Apple Inc., which is the stock symbol used to represent ... 
        \tcblower        
        \begin{center}
            Examples of Wrong Output
        \end{center}
        \textbf{Llama 2 7B}: \\
        \textbf{Prompt}: In finance, the acronym "PPI" stands for \\
        \textbf{Generation}: Price To Book Ratio. \\
        \textbf{Prompt}: In finance, the stock symbol "ZVIA" stands for \\
        \textbf{Generation}: "Zions Bancorporation". \\
        
        \textbf{Llama 2 7B Chat}: \\
        \textbf{Prompt}: In finance, what does the acronym "PPI" stand for? \\
        \textbf{Generation}: Personal Protection Insurance. \\
        \textbf{Prompt}: In finance, what does the stock ticker "ZVIA" stand for? \\
        \textbf{Generation}: "ZVIA" stands for Zvia Food \& Beverage, Inc. 
    \end{tcolorbox}
    \caption{Examples of zero-shot output for the Llama2 models on the two abbreviation tasks. }
    \label{tab:app_abbr_examples}
\end{table}

\begin{table}[!ht]
    \centering
    \begin{tcolorbox}
        \textbf{Llama 2 7B}: \\
        \textbf{Prompt}: In the context of finance, explain the concept of "Security Commission.\textbackslash n\textbackslash n Security Commission\\
        \textbf{Generation}: is a regulatory body that monitors the financial market in India. It was established in 1987 to regulate the capital markets and protect investors' interests. The commission has several functions, including:\textbackslash n\textbackslash n\* Regulating the securities market by issuing rules and regulations governing the activities of brokers, traders, and other participants in the market.\textbackslash n\textbackslash n\* Monitoring compliance with these rules and regulations.\textbackslash n\textbackslash n* Taking action against violators.\textbackslash n\textbackslash n* Promoting transparency and fairness in the market.\textbackslash n\textbackslash nThe Security Commission also provides guidance on best practices for investment and risk management. This includes providing educational materials and hosting seminars and workshops to help investors make informed decisions about their investments. \\

        \textbf{Llama 2 7B Chat}: \\
        \textbf{Prompt}: In the context of finance, explain the concept of "Security Commission".\\ 
        \textbf{Generation}: Security Commission is a term used in the financial industry to refer to an organization that oversees and regulates the securities market. The main function of a Security Commission is to protect investors by ensuring that securities firms operate fairly and honestly, and that they comply with relevant laws and regulations.\textbackslash nIn many countries, the Security Commission is responsible for issuing licenses to securities firms, approving their financial statements, and monitoring their activities to ensure that they are acting in the best interests of their clients. The Security Commission may also investigate and take action against firms that violate securities laws or engage in unethical practices.\textbackslash nThe Security Commission plays a critical role in maintaining the integrity of the securities markets and protecting investors from fraudulent activities. By setting and enforcing strict standards for securities firms, the Security Commission helps to ensure that investors have access to a fair and transparent marketplace where they can buy and sell securities with confidence. \\

    \end{tcolorbox}
    \caption{Examples of zero-shot output for the Llama2 models on the finance terminology task. }
    \label{tab:app_term}
\end{table}

\begin{table}[!ht]
    \centering
    \begin{tcolorbox}
        \begin{center}
            Vanilla Zero-Shot
        \end{center}
        \textbf{Llama 2 7B}: \\
        \textbf{Prompt}: The highest stock price for ticker symbol "CC" on the date of 2022-05-23 was \\
        \textbf{Generation}: 1. nobody knows what will happen in the future, but it is a good idea to be prepared for any scenario.\textbackslash nThe stock price of 'CC' has been on a downward trend over the past year and has recently fallen below its ...\\

        \textbf{Llama 2 7B Chat}: \\
        \textbf{Prompt}: What was the highest stock price for ticker symbol "CC" on the date of 2022-05-23?\\
        \textbf{Generation}: The highest stock price for ticker symbol "CC" on the date of 2022-05-23 was \$149.99.\\
        \tcblower        
        \begin{center}
            Zero-Shot Tool Learning
        \end{center}
        \textbf{Prompt}: \\
        Instruction:\\
        For each natural language question below, generate the corresponding Python function call using the get\_stock\_price\_on\_date(ticker, date, price\_type) function. \\
        Documentation: \\
        Use the ticker parameter for the stock's ticker symbol mentioned in the question.\\
        Use the date parameter for the specified date in the question.\\
        The price\_type parameter should be either "highest" or "lowest". \\
        \\Question:\\
        What was the highest stock price for ticker symbol `CC' on the date of 2023-09-22?\\
        Answer of the Python function call:\\\\
        \textbf{Generation}: \textbackslash n```python\textbackslash nget\_stock\_price\_on\_date(`CC', `2023-09-22', `highest')\\

        \textbf{Llama 2 7B Chat}: \\
        \textbf{Prompt}: \\
        Instruction:\\
        For each natural language question below, generate the corresponding Python function call using the get\_stock\_price\_on\_date(ticker, date, price\_type) function. \\
        Documentation: \\
        Use the ticker parameter for the stock's ticker symbol mentioned in the question.\\
        Use the date parameter for the specified date in the question.\\
        The price\_type parameter should be either "highest" or "lowest". \\
        \\Question:\\
        What was the highest stock price for ticker symbol `CC' on the date of 2023-09-22?\\
        Answer of the Python function call:\\\\
        \textbf{Generation}: get\_stock\_price\_on\_date(`CC', `2023-09-22', 'highest')\\
    \end{tcolorbox}
    \caption{Examples of vanilla zero-shot and zero-shot tool learning output for the Llama2 models on the stock price query task. }
    \label{tab:app_price}
\end{table}

\clearpage
\section{Few-Shot Prompt Templates}
\begin{table}[h!]
    \centering
    \begin{tcolorbox}[width=\linewidth]
        \begin{center}
            Financial Acronym Task
        \end{center}
        Question: In finance, what does the acronym "IPO" stand for? \\
        Answer: Initial Public Offering.\\
        \\
        Question: In finance, what does the acronym "ROI" stand for?\\
        Answer: Return on Investment.\\
        \\
        Question: In finance, what does the acronym "ETF" stand for?\\
        Answer: Exchange-Traded Fund.\\
        \\
        Question: In finance, what does the acronym "ESG" stand for?\\
        Answer: Environmental, Social, and Governance.\\
        \\
        Question: In finance, what does the acronym "APR" stand for?\\
        Answer: Annual Percentage Rate.\\
        \\
        Question: In finance, what does the acronym "[test acronym]" stand for?\\
        Answer: \\
        \tcblower        
        \begin{center}
            Stock Symbol Task
        \end{center}
        Question: In finance, what is the company name for the stock ticker "ADA"?\\
        Answer: The company name for the stock symbol "ADA" is Adams Plc.\\
        \\
        Question: In finance, what is the company name for the stock symbol "GBG"?\\
        Answer: The company name for the stock symbol "GBG" is Gb Group Plc.\\
        \\
        Question: In finance, what is the company name for the stock symbol "LGRS"?\\
        Answer: The company name for the stock symbol "LGRS" is Loungers Plc.\\
        \\
        Question: In finance, what is the company name for the stock symbol "AIG"?\\
        Answer: The company name for the stock symbol "AIG" is American International Group.\\
        \\
        Question: In finance, what is the company name for the stock symbol "RGR"?\\
        Answer: The company name for the stock symbol "RGR" is Sturm Ruger \& Company.\\
        \\
        Question: In finance, what is the company name for the stock symbol "MAIN"?\\
        Answer: The company name for the stock symbol "MAIN" is Main Street Capital Corp. \\
        \\
        Question: In finance, what is the company name for the stock symbol "[stock symbol]"?\\
        Answer: \\
    \end{tcolorbox}
    \caption{The few-shot demonstration templates used in the financial two abbreviation task.}
    \label{tab:demo_abbr}
\end{table}

\begin{table}[ht!]
\begin{tcolorbox}
Question: In the context of finance, explain the concept of "Compound Interest".\\
Answer: Compound interest is the interest calculated on the initial principal, which also includes all the accumulated interest from previous periods on a deposit or loan. Thought of as "interest on interest," it can significantly increase savings or costs over time, as the interest is calculated on a cumulative basis. Unlike simple interest, which is calculated only on the principal amount, compound interest is calculated on the principal amount plus any interest that's been added previously.\\
\\
Question: In the context of finance, explain the concept of "Hedge Funds".\\
Answer: Hedge funds are alternative investments using pooled funds that employ different strategies to earn active return, or alpha, for their investors. Hedge funds may use a wide range of strategies, including short selling, leverage, swaps, derivatives, and arbitrage. They are typically open to a limited range of investors and require a large initial minimum investment. Hedge funds are known for their aggressive investment strategies aimed at yielding high returns.\\
\\
Question: In the context of finance, explain the concept of "Equity".\\
Answer: In finance, equity refers to the ownership of assets after all debts associated with those assets are paid off. For example, in stock trading, equity represents the ownership interest held by shareholders in a corporation, measured in the unit of shares. In the context of real estate, equity means the difference between the property's current market value and the amount the owner still owes on the mortgage. It represents the amount that the property owner would receive after selling the asset and paying off the mortgage.\\
\\
Question: In the context of finance, explain the concept of "[term]".\\
Answer: 
\end{tcolorbox}
\caption{The few-shot demonstration template used in the financial terminology explanation task. }
\label{tab:demo_long-form}
\end{table}

\begin{table}[!htbp]
\small
\centering
\begin{tcolorbox}
\begin{center}
Vanilla Few-Shot
\end{center}
Question: What was the highest stock price for stock symbol "QS" on the date of 2022-05-23?\\
Answer: \$12.1300.\\
\\
Question: What was the lowest stock price for stock symbol "QS" on the date of 2022-05-23?\\
Answer: \$11.2750.\\
\\
Question: What was the highest stock price for stock symbol "QS" on the date of 2022-06-22?\\
Answer: \$9.5400.\\
\\
Question: What was the lowest stock price for stock symbol "QS" on the date of 2022-06-22?\\
Answer: \$8.8100.\\
\\
Question: What was the highest stock price for stock symbol "QS" on the date of 2022-07-22?\\
Answer: \$12.0900.\\
\\
Question: What was the lowest stock price for stock symbol "QS" on the date of 2022-07-22?\\
Answer: \$10.8400.\\
\\
Question: What was the highest stock price for stock symbol "QS" on the date of 2022-08-22?\\
Answer: \$11.1391.\\
\\
Question: What was the lowest stock price for stock symbol "QS" on the date of 2022-08-22?\\
Answer: \$10.6600.\\
\\
Question: What was the [lowest/highest] stock price for stock symbol "[stock symbol]" on the date of [date]?\\
Answer: 
\tcblower        
\begin{center}
One-Shot Tool Learning
\end{center}
    Instruction:\\
    For each natural language question below, generate the corresponding Python function call using the get\_stock\_price\_on\_date(ticker, date, price\_type) function. \\
    Documentation: \\
    Use the ticker parameter for the stock's ticker symbol mentioned in the question.\\
    Use the date parameter for the specified date in the question.\\
    The price\_type parameter should be either "highest" or "lowest". \\
    \\Question:\\
    What was the highest stock price for ticker symbol `CC' on the date of 2023-09-22?\\
    Answer of the Python function call: \\
    get\_stock\_price\_on\_date(`CC', `2023-09-22', `highest')\\
    \\Question:\\
    What was the [lowest/highest] stock price for stock symbol "[stock symbol]" on the date of [date]?\\
    Answer of the Python function call:\\

\end{tcolorbox}
\caption{The few-shot demonstration template used in the stock price query task. }
\label{tab:demo_price}
\end{table}

\end{document}